\documentclass[sigconf,authorversion,nonacm]{acmart}

\AtBeginDocument{%
  \providecommand\BibTeX{{%
    \normalfont B\kern-0.5em{\scshape i\kern-0.25em b}\kern-0.8em\TeX}}}

\PassOptionsToPackage{dvipsnames,prologue, table,x11names}{xcolor}
\usepackage{todonotes}
\usepackage{times}
\usepackage{latexsym}
\usepackage{amsmath}
\usepackage{url}
\usepackage{color, colortbl}
\usepackage{multirow}
\usepackage{graphicx}
\usepackage{graphics}
\usepackage{enumitem}
\usepackage{subcaption}

\usepackage{amssymb}
\usepackage[normalem]{ulem} 
\usepackage{makecell}
\usepackage{array}
\usepackage{float}
\usepackage{breqn}
\usepackage{arydshln}

\usepackage{listings}
\usepackage[lined,ruled]{algorithm2e}

\usepackage{siunitx}

\definecolor{ForestGreen}{rgb}{0.0, 0.4, 0.0}
\definecolor{airforceblue}{rgb}{0.36, 0.74, 0.66}
\definecolor{blue(ncs)}{rgb}{0, 0.33, 0.5}
\definecolor{bleudefrance}{rgb}{0.19, 0.55, 0.91}
\definecolor{cobalt}{rgb}{0.0, 0.28, 0.67}
\definecolor{Gray}{gray}{0.9}

\newlength{\Oldarrayrulewidth}
\newcommand{\Cline}[2]{%
  \noalign{\global\setlength{\Oldarrayrulewidth}{\arrayrulewidth}}%
  \noalign{\global\setlength{\arrayrulewidth}{#1}}\cline{#2}%
  \noalign{\global\setlength{\arrayrulewidth}{\Oldarrayrulewidth}}}

\acmConference[]{}{}

\acmSubmissionID{1202}

\begin{document}

\title{Unified Generative \& Dense Retrieval for Query Rewriting in Sponsored Search}

\author{Akash Kumar Mohankumar}
\email{makashkumar@microsoft.com}
\affiliation{%
  \institution{Microsoft}
  \city{Bengaluru}
  \country{India}
}
  
\author{Bhargav Dodla}
\authornote{Work done during internship at Microsoft}
\email{dodla.bhargav@gmail.com}
\orcid{1234-5678-9012}
\affiliation{%
  \institution{Indian Institute of Technology Madras}
  \city{Chennai}
  \country{India}
}

\author{Gururaj K}
\email{gururajk@microsoft.com}
\affiliation{%
  \institution{Microsoft}
  \city{Bengaluru}
  \country{India}
}

\author{Amit Singh}
\email{siamit@microsoft.com}
\affiliation{%
  \institution{Microsoft}
  \city{Bengaluru}
  \country{India}
  }








\begin{abstract}
Sponsored search is a key revenue source for search engines, where advertisers bid on keywords to target users or search queries of interest. However, finding relevant keywords for a given query is challenging due to the large and dynamic keyword space, ambiguous user/advertiser intents, and diverse possible topics and languages. In this work, we present a comprehensive comparison between two paradigms for online query rewriting: Generative (NLG) and Dense Retrieval (DR) methods. We observe that both methods offer complementary benefits that are additive. As a result, we show that around 40\% of the high-quality keywords retrieved by the two approaches are unique and not retrieved by the other. To leverage the strengths of both methods, we propose CLOVER-Unity, a novel approach that unifies generative and dense retrieval methods in one single model. Through offline experiments, we show that the NLG and DR components of CLOVER-Unity consistently outperform individually trained NLG and DR models on public and internal benchmarks. Furthermore, we show that CLOVER-Unity achieves 9.8\% higher good keyword density than the ensemble of two separate DR and NLG models while reducing computational costs by almost half. We conduct extensive online A/B experiments on Microsoft Bing in 140+ countries and achieve improved user engagement, with an average increase in total clicks by 0.89\% and increased revenue by 1.27\%. We also share our practical lessons and optimization tricks for deploying such unified models in production. 
\end{abstract}


\keywords{sponsored search, dense retrieval, natural language generation}


\maketitle
\section{Introduction}
\begin{table*}[h]
    \centering
	\begin{center}
	 \begin{subtable}[h]{\textwidth}
        \centering
		\small\addtolength{\tabcolsep}{-4pt}
		\scalebox{0.95}{
		\begin{tabular}{c  c  c  c }	
			\Xhline{3\arrayrulewidth}
			{Query} & {Generative Retrieval} & {Dense Retrieval}  & {CLOVER-Unity} \\ \hline
			& black 912	& currys {\textcolor{red}{kettles}}  & {\textcolor{ForestGreen}{hp}} 912 {\textcolor{ForestGreen}{ink cartridges}} black\\
			{\textcolor{blue(ncs)}{currys 912 black}}&currys currys&  currys {\textcolor{red}{electrical kettles}} & currys {\textcolor{ForestGreen}{ink cartridges}} \\ 
			& {\textcolor{red}{black curry}}  &  {\textcolor{red}{kettles}} at currys 	& black {\textcolor{ForestGreen}{hp printer cartridges}} \\ \hline
      	\multirow{3}{*}{\begin{tabular}[c]{@{}c@{}}{\textcolor{blue(ncs)}{student houses for rent}}\\ {\textcolor{blue(ncs)}{nottingham}} \end{tabular}} & student {\textcolor{ForestGreen}{properties}} nottingham & rentals in nottinghamshire {\textcolor{ForestGreen}{uk}} & nottingham student {\textcolor{ForestGreen}{studios}} \\
			 & student {\textcolor{ForestGreen}{accom}} nottingham & rentals nottingham {\textcolor{ForestGreen}{united kingdom}} & student {\textcolor{ForestGreen}{flats}} in nottingham \\ 
			&rent house rental & {\textcolor{red}{3 bedroom}} house nottingham	& student {\textcolor{ForestGreen}{lettings}} nottingham {\textcolor{ForestGreen}{united kingdom}}\\ \hline

			&250 honda & honda  & honda 250 {\textcolor{ForestGreen}{cc motorcycle}} \\
			{\textcolor{blue(ncs)}{06 honda 250}}& honda motor  & hondas 250 &  06 honda {\textcolor{ForestGreen}{cbr}} 250\\ 
			&honda250 & {\textcolor{red}{car for lease}} & honda {\textcolor{ForestGreen}{2006}} bike \\ \hline

			\multirow{6}{*}{\begin{tabular}[c]{@{}c@{}}  {\textcolor{blue(ncs)}{cappella sistina biglietti}}\\ {(\textcolor{blue(ncs)}{sistine chapel tickets})} \end{tabular}}
			& \multirow{2}{*}{\begin{tabular}[c]{@{}c@{}} biglietti per la cappella sistina	\\ {(tickets for the sistine chapel)}	\end{tabular}}
			& \multirow{2}{*}{\begin{tabular}[c]{@{}c@{}} sistina roma\\ {(sistine rome)}	\end{tabular}}
			& \multirow{2}{*}{\begin{tabular}[c]{@{}c@{}} cappella sistina excursions\\ {(sistine chapel {\textcolor{ForestGreen}{excursions}})}\end{tabular}} \\ \\ \cline{2-4}
			& \multirow{2}{*}{\begin{tabular}[c]{@{}c@{}} capella sistina	\\ (sistine chapel)	\end{tabular}}
			& \multirow{2}{*}{\begin{tabular}[c]{@{}c@{}} cappella sistina museum \\ (sistine chapel museum)	\end{tabular}}
			& \multirow{2}{*}{\begin{tabular}[c]{@{}c@{}} cappella sistina vatican city \\ (sistine chapel {\textcolor{ForestGreen}{vatican city}})\end{tabular}} \\ \\ \cline{2-4}
			& \multirow{2}{*}{\begin{tabular}[c]{@{}c@{}} la cappella \\ (the chapel)	\end{tabular}}
			& \multirow{2}{*}{\begin{tabular}[c]{@{}c@{}} cappella sistina orari e prezzi \\ (sistine chapel \textcolor{red}{times and prices})	\end{tabular}}
			& \multirow{2}{*}{\begin{tabular}[c]{@{}c@{}} musei vaticani e cappella sistina prenotazione \\ ({\textcolor{ForestGreen}{vatican museums}} and sistine chapel booking)\end{tabular}} \\ \\
			\Xhline{3\arrayrulewidth}
		\end{tabular}	
		}
		\end{subtable}
			\caption{Comparison of keywords retrieved by CLOVERv2, NGAME and our proposed CLOVER-Unity approach for four queries}
		\label{tab:Example}	
	\end{center}
	\end{table*}
\textbf{Overview:} 
Sponsored search is a form of online advertising that enables advertisers to display their ads along with organic results on web search engines. It is a major source of revenue for search engines and helps advertisers attract targeted user traffic. To participate in sponsored search, advertisers can bid on keywords that are relevant to their products or services using different match types. For instance, an exact match bid keyword will only match search queries that have the same search intent as the keyword. Conversely, phrase match will match queries that contain or include the meaning of the keyword. Matching search queries with relevant keywords, also referred to as query rewriting, is a complex and challenging task for various reasons. First, search queries and bid keywords are often short and ambiguous, making it difficult to precisely infer the user’s or advertiser’s intent. For example, the query \textit{"currys 912 black"} is vague, but the user is actually looking for the HP 912 ink cartridge in black color from the retailer Currys. Second, search queries and keywords can span a wide range of topics, domains, languages, and countries, adding to the complexity of the task. Third, a search query must be matched with all possible high-quality exact/phrase keywords, not just one, from a large collection of bid keywords. Lastly, the matching must be done in real-time and be computationally efficient to handle all search traffic. \\

\textbf{Existing methods} for query rewriting matching can be broadly categorized into two main groups: information retrieval (IR) and generative or NLG based retrieval. IR methods use various techniques to learn sparse or dense representations for queries and keywords based on bag-of-words, static pretrained features, or deep learning models \cite{Broder2007ASA, Broder2008SearchAU, RibeiroNeto2005ImpedanceCI, Prabhu2020ExtremeRF, Prabhu2018ExtremeML, parabel, Bai2018ScalableQN, decaf, SiameseXML}. Among them, recent Dense Retrieval (DR) methods that leverage effective negative mining strategies such as ANCE \cite{ANCE}, RocketQA \cite{RocketQA}, NGAME \cite{NGAME} have been shown to achieve state-of-the-art performance on various retrieval benchmarks \cite{manik_xc_repo}. On the other hand, NLG-based methods for query rewriting such as CLOVER \cite{cloverv1}, ProphetNet-Ads \cite{ProphetNetAds} use generative models to directly transform queries into keywords. This involves training language models to generate query rewrites and then constraining their generation space during inference to the set of bid keywords. \\

Given these two distinct approaches to the same problem, we perform a comprehensive study comparing their performance. Our results show NLG and DR methods retrieve a similar number of high-quality keywords per query, but many of them are different and non-overlapping. Specifically, about 40\% of the high-quality keywords retrieved by DR and NLG are unique to each method and weren't retrieved by the other. We analyze the reasons for these and find they stem from the structural differences between the approaches. For instance, NLG methods treat query rewriting as a token-level generation task with a token-level loss and thus  are better at learning word-level relationships such as synonymy and hypernymy. As shown in example 2 of Table \ref{tab:Example}, NLG models can better identify interchangeable tokens, such as \textit{houses} and \textit{properties} or \textit{accom}, which do not alter the query/keyword intent. However, we also find that NLG methods, especially the non-autoregressive ones, struggle to retrieve longer keywords that may have a few additional tokens. This is because non-autoregressive NLG models have to predict all $n$ tokens of the keyword in parallel during inference. In contrast, DR methods do not depend on keyword token length during inference and thus do not have this limitation. In the same example in Table \ref{tab:Example}, DR methods retrieve keywords that contain additional tokens such as \textit{united kingdom}, which NLG models miss. \\

With NLG and DR models retrieving substantially different high-quality keywords for the same query, it leads to a natural question of whether we can combine the strengths of both approaches. To this end, we propose CLOVER-Unity, a unified retriever that blends the advantages of NLG and DR methods using a shared encoder. Unlike standard multi-task learning, where a shared model solves more than one related task, CLOVER-Unity solves the same task (query rewriting) using two different methods: NLG and DR. Through offline experiments, we show that the NLG and DR components of CLOVER-Unity consistently outperform the individually trained NLG and DR models by an average of 19.2\% and 7.4\%, respectively, in terms of good keyword density. We attribute these improvements to training the Unity model with both the contrastive DR loss and the token-level NLG loss. For instance, we show that the DR component of CLOVER-Unity is better at identifying token-level relationships due to the additional token-level objective function. Further, we show that a single CLOVER-Unity model with one forward pass retrieves 9.8\% more good keywords than the ensemble of two separate DR and NLG models while using 45.8\% lower GPU compute. We also conduct extensive online experiments on Microsoft Bing in 140+ countries and achieve significant gains in revenue, clicks, impressions, and coverage, demonstrating the practical utility of CLOVER-Unity for query rewriting in sponsored search. To the best of our knowledge, we are the first to present such a unified framework for query rewriting. We also share our practical lessons and optimization tricks for deploying such unified models in production. Finally, we demonstrate the versatility of CLOVER-Unity by applying it to problems beyond sponsored search. We evaluate it on public datasets from the extreme classification repository \cite{manik_xc_repo} and show that CLOVER-Unity outperforms the leading extreme classification algorithms in all metrics.
\section{Proposed Method}
\begin{figure}
    \centering
    \includegraphics[width=0.45\textwidth]{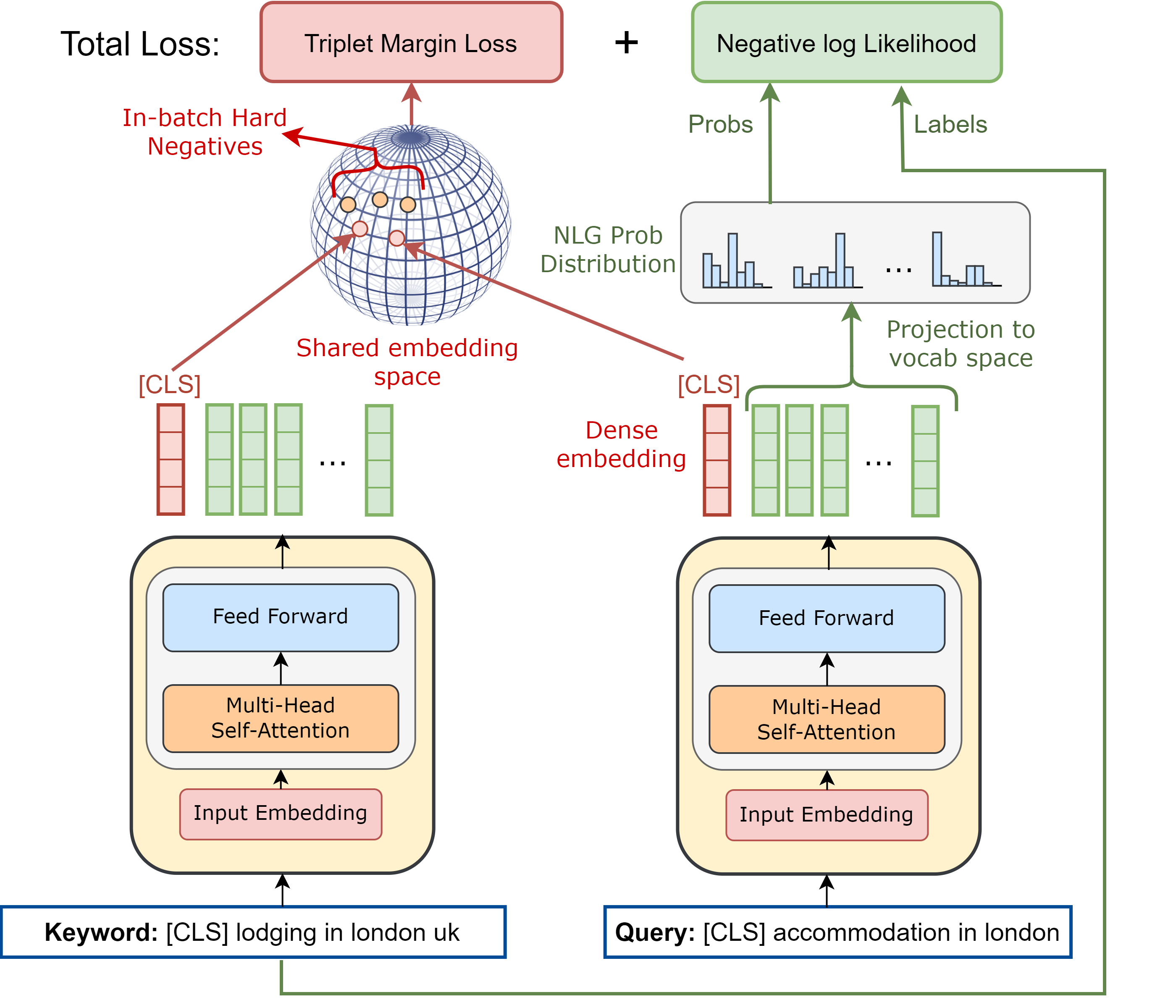}
    \caption{Training process of CLOVER-Unity using contrastive DR and log-likelihood NLG loss functions}
    \label{fig:unity_model}
\end{figure}
In this section, we describe the various components of our proposed approach for query rewriting. Our goal is to match any given search query $q$ with all its relevant exact bid keywords $k^{e}$ and phrase bid keywords $k^{p}$ in real-time. We first present details of the NLG and DR baselines in \ref{subsec:proposed_generative} and \ref{subsec:proposed_dense}. Then, we introduce CLOVER-Unity, which combines the benefits of both methods in a single model.
\subsection{Generative Retrieval}
\label{subsec:proposed_generative}
NLG models treat query rewriting as a token-level prediction task, \textit{i.e.}, they predict relevant keyword tokens $\{k_t\}_{t=1}^{m}$ individually instead of retrieving them directly. However, predicting tokens one at a time in an autoregressive (AR) fashion is very slow as it requires $m$ sequential forward passes over the model ($m$ is the keyword length). Moreover, when performing beam search with beam size $B$, each forward pass requires $O(B)$ compute. As we show in section \ref{subsec:generative_vs_dense}, existing AR models like CLOVER \cite{cloverv1}, ProphetNetAds \cite{ProphetNetAds}, and even efficient variants with deep encoder and shallow decoder \cite{deepshallow} have an order of magnitude higher inference cost than dense retrieval approaches. To overcome these issues, we propose CLOVERv2, which uses an encoder-based non-autoregressive (NAR) model with AR trie decoding for efficient inference. Specifically, we first assume the keyword tokens $K_t$ are conditionally independent given the query, \textit{i.e.}, the probability distribution can be factorized as:
$$
    p^{NAR}(K_1, \dots, K_m | Q, \theta) = \prod_{t=1}^{m} p^{NAR}(K_t | Q, \theta)
$$
This decomposition allows us to predict the token distributions independently in parallel, using only one forward pass through an encoder model such as BERT \cite{bert} with a language modeling head.

{\flushleft \textbf{Decoding:}} We construct a trie $T_{\mathcal{K}}$ of valid bid keywords $\mathcal{K}$ to constrain the generation space to the closed set $\mathcal{K}$. We developed an optimized C++ implementation of the trie data structure that is highly memory efficient and can be hosted even on commodity CPU machines. For example, a trie of 1 billion keywords (45 GB of raw text) requires only 6.5 GB of CPU RAM. The trie plays a crucial role in decoding; it functions as an autoregressive language model to guide the generation process. In particular, we proceed in a left-to-right manner, where at each time step $t$, we consider only the children of the partially generated prefix $k_{<t}$ in $T_{\mathcal{K}}$ as possible next tokens. Essentially, decoding a NAR model with trie is equivalent to an AR language model with (unnormalized) probability distribution:
\begin{equation*}
    \tilde{p}(K_t=k_t | k_{<t}, q, {\mathcal{K}}) =     
\begin{cases}
    p^{NAR}(K_t=k_t | q),& \text{if } k_t\in \text{child}_{T_{\mathcal{K}}}(k_{<t})\\
    0,              & \text{otherwise}
\end{cases}
\end{equation*}
We note that NAR models with AR trie decoding (i) do not suffer from issues such as token repetition or mode mixing observed in vanilla NAR \cite{NAR},  and (ii) achieve comparable or better performance than AR models while being more efficient (section \ref{subsec:generative_vs_dense}).   

\subsection{Dense Retrieval}
\label{subsec:proposed_dense}
 We adopt a siamese architecture with shared parameters to learn dense representations for queries and keywords. To train the encoder effectively, we need to carefully select a small set of irrelevant keywords for each query. We cannot use all irrelevant keywords for a query because the number of training points and bid keywords is very large (millions or billions). Moreover, randomly choosing negative keywords from a uniform distribution or from the same batch can result in uninformative negatives and slow convergence \cite{ANCE}. To address these issues, we use NGAME \cite{NGAME}, a negative mining technique that selects hard negative samples within the training batches. NGAME has been shown to outperform other approaches such as ANCE \cite{ANCE}, RocketQA \cite{RocketQA}, and TAS \cite{TAS}. In our application, we omit the classifier layer in NGAME because bid keywords $\mathcal{K}$ change over time. After training, we index the keyword embeddings on SSD using the DiskANN algorithm \cite{DiskANN}, which requires minimal CPU RAM online. During inference, we retrieve the keywords that are approximately the top k nearest neighbors of the query embedding. 
\subsection{Unified Retrieval}
\label{subsec:proposed_unified}
We now introduce CLOVER-unity, which uses a variant of the standard multi-task learning setup where we solve one task with two different methods. In particular, we use a single encoder to perform dense and generative retrieval simultaneously, as shown in Figure \ref{fig:unity_model}. Given a query $q = \{q_1, \dots, q_n\}$, we prepend a special [CLS] token and encode the query using a transformer encoder $\mathcal{E}_{\theta}$. The resulting sequence of hidden states, $h^q = \{h^q_c, h^q_1, \dots, h^q_{n}\}$, is used to compute the DR and NLG representations of the query. The first hidden state, $h^q_c$, serves as the DR representation, while the remaining hidden states, $h^q_t \in \mathbb{R}^{d}$, are projected to the output vocabulary space to obtain the NAR logits, $\log \Tilde{p}(K_t | q, \theta) = \mathbf{W}h^q_t$ where $\mathbf{W} \in \mathbb{R}^{V \times d}$ is a learnable weight matrix, $V$, and $d$ are vocabulary and hidden sizes. We apply the same process to keywords, obtaining a DR representation $h^k_c$ for each keyword. 

{\flushleft \textbf{Training:}} We train CLOVER-Unity on a supervised dataset $\mathcal{D} = \{(q^{(i)}, k^{(i)})\}_{i=1}^{L}$ of high-quality query keyword pairs. For each query $q^{(i)}$, we obtain hard negative keywords $l^{(i)}$ using the NGAME negative mining technique. Our objective is to minimize a weighted combination of the triplet margin loss for DR and negative log-likelihood for NLG, as defined by the following equation:
\begin{align*}
    \mathcal{L}(\theta, \mathcal{D}) =   \frac{1}{L} \sum_{q, k \in \mathcal{D}} \Bigg( \Bigg. \left[ (h^{l}_c)^{T}h^{q}_c - (h^{k}_c)^{T}h^{q}_c + \gamma \right]_{+} \\
    - \alpha \sum_{t=1}^{m} \log p(K_t = k_t | q, \theta) \Bigg. \Bigg)
\end{align*}
where $h^q_c$, $h^k_c$, and $h^l_c$ are the DR representations of the query, relevant, and irrelevant keywords, respectively.  
{\flushleft \textbf{Inference:}} During inference, we obtain the query representation $h^q_c$ and the NAR probabilities $p(K_t | q, \theta)$ through a single pass over the encoder $\mathcal{E}_{\theta}$. We then simultaneously retrieve bid keywords using (i) DiskANN search on the query vector $h^q_c$, and (ii) trie-based beam search on the predicted probabilities $p(K_t | q, \theta)$. However, we observed that using standard beam search for decoding NLG models tends to produce very similar keywords with common prefixes and low diversity. To address this issue, we propose permutation decoding, which leverages a property of NAR models: they do not have an explicit order of generation. Therefore, we decode NAR models in multiple orders (e.g., left-to-right and right-to-left) and rank the top B results based on cumulative log probability, which remains unchanged regardless of the decoding order.
\section{Offline Experiments}
\subsection{Experimental Setup}
\begin{table}[]
\centering
\begingroup
\def\arraystretch{1.19}
\resizebox{0.98\linewidth}{!}{
\begin{tabular}{lcccc}
\Xhline{2\arrayrulewidth}
\multicolumn{1}{l|}{Dataset}                & \multicolumn{1}{c|}{Labels}     & \multicolumn{1}{c|}{\begin{tabular}[c]{@{}c@{}}\#Train \\ Data Points\end{tabular}} & \multicolumn{1}{c|}{\begin{tabular}[c]{@{}c@{}}\#Test \\ Data Points\end{tabular}} & \begin{tabular}[c]{@{}c@{}}Avg. Labels\\ per  point\end{tabular} \\ \hline
\multicolumn{5}{c}{Sponsored Search}                                                                                                                                                                                                                                                                                       \\ \hline
\multicolumn{1}{l|}{Query2Bid-1M}           & \multicolumn{1}{c|}{997,994}    & \multicolumn{1}{c|}{16,366,540}                                                     & \multicolumn{1}{c|}{500,000}                                                       & 1.68                                                             \\ \hline
\multicolumn{1}{l|}{Query2Bid-5M}           & \multicolumn{1}{c|}{4,995,652}  & \multicolumn{1}{c|}{49,571,313}                                                     & \multicolumn{1}{c|}{500,000}                                                       & 2.80                                                             \\ \hline
\multicolumn{1}{l|}{Query2Bid-10M}          & \multicolumn{1}{c|}{10,011,695} & \multicolumn{1}{c|}{75,532,741}                                                     & \multicolumn{1}{c|}{500,000}                                                       & 3.82                                                             \\ \hline
\multicolumn{5}{c}{Public Short-text Benchmarks}                                                                                                                                                                                                                                                                           \\ \hline
\multicolumn{1}{l}{AmazonTitles}      & \multicolumn{1}{c|}{131,073}    & \multicolumn{1}{c|}{294,805}                                                        & \multicolumn{1}{c|}{134,835}                                                       & 2.29                                                             \\ \hline
\multicolumn{1}{l}{WikiSeeAlsoTitles} & \multicolumn{1}{c|}{312,330}    & \multicolumn{1}{c|}{693,082}                                                        & \multicolumn{1}{c|}{177,515}                                                       & 2.11                                                             \\ \Xhline{2\arrayrulewidth}
\end{tabular}}
\caption{Summary of datasets used for offline analysis, with labels indicating bid keywords for the sponsored search datasets}
\label{tab:dataset_stats}
\endgroup
\end{table}
{\flushleft \textbf{Datasets:}} We curated three sponsored search datasets from Bing Search logs spanning 30+ languages across all geographies: Query2Bid-1M, Query2Bid-5M, and Query2Bid-10M. These datasets consist of high-quality query-keyword pairs mined from various sources such as ad impression logs and offline dictionaries. To demonstrate the versatility of our proposed approach, we also included two widely used short-text retrieval benchmarks from the extreme classification repository \cite{manik_xc_repo}: AmazonTitles and WikiSeeAlsoTitles. Table \ref{tab:dataset_stats} shows the statistics of these datasets.

{\flushleft \textbf{Evaluation Metrics:}} Query rewriting models are challenging to evaluate. Standard metrics such as Precision and Recall are biased and unreliable due to missing ground truths in the large keyword space. Moreover, we found that existing NLG evaluation metrics such as BLEU \cite{bleu}, BertScore \cite{bert_score}, and Bleurt \cite{bleurt} correlate poorly with human judgments with Pearson’s r less than 0.2. Therefore, following \cite{cloverv1}, we trained a custom evaluation model on a large-scale human-annotated dataset. The trained model provides a separate score for exact and phrase match quality, denoted by $\mathcal{M}^{e}(q,k)$ and $\mathcal{M}^{p}(q,k)$, for any given query-keyword pair $q,k$. It achieves an AUC-ROC score of $0.837$ with binary human judgments, outperforming existing NLG metrics by ~35\%. We use this evaluation model to compute the Good Keyword Density measure $Q$, defined as the average number of high-quality keywords retrieved per query:
\begin{align*}
    \mbox{Exact } Q@s &= \frac{|\{q,k: \mathcal{M}^{e}(q,k) > s\}|}{n}\\
    \mbox{Phrase } Q@s &= \frac{|\{q,k: \mathcal{M}^{p}(q,k) > s\}|}{n}
\end{align*}
where $\{q_i, k_i\}_{i=1}^{m}$ is the retrieved set, $n$ is the number of test queries, and $s$ is a threshold applied to the evaluation scores. We report exact and phrase $Q@0.5$ and $Q@0.7$ in all our experiments. For the public retrieval datasets, we use the standard evaluation measures \cite{xc_metrics}: Precision$@k$ (P$@k$),  nDCG$@k$ (N$@k$), propensity-scored Precision (PSP$@k$) and nDCG (PSN$@k$) and Recall$@k$ (R$@k$). 
{\flushleft \textbf{Implementation Details:}} We initialize CLOVER-Unity and other baselines with the pre-trained XLM-R base model \cite{xlmr} and finetune on 4x Nvidia 40GB A100 GPUs with max sequence length of 16 and batch size of 1024. We use a beam size of 50 for AR and 100 for NAR NLG models (unless specified). For DR, we fetch the top 100 keywords during ANN search. We compute the cost to serve models on Nvidia T4 GPUs using the price calculator on Azure cloud. 
\subsection{Generative vs Dense Retrieval}
\begin{figure}
    \centering
    \includegraphics[width=0.45\textwidth]{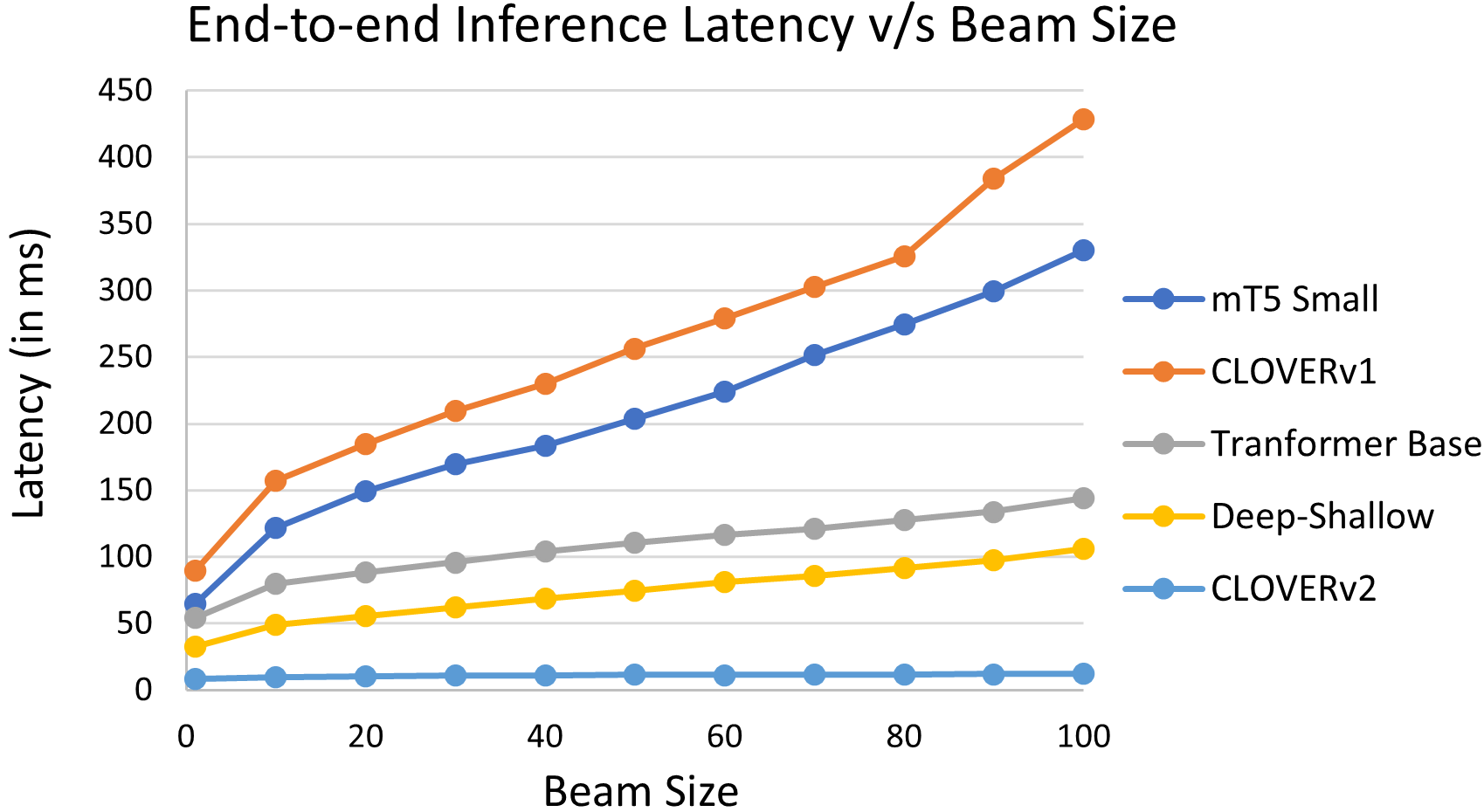}
    \caption{End-to-end inference latency (in ms) of AR and NAR NLG models for different beam sizes}
    \label{fig:latency_beam}
\end{figure}
\label{subsec:generative_vs_dense}
\begin{table*}[]
    \centering
    \begingroup
    \def\arraystretch{1.19}
    \resizebox{0.96\textwidth}{!}{
    \begin{tabular}{cl|cccc|cccc|cccc|c|c}
    \Cline{2pt}{2-16}
                                   &  & \multicolumn{4}{c|}{\textbf{Query2Bid-1M}}                                                                                                                                                                    & \multicolumn{4}{c|}{\textbf{Query2Bid-5M}}                                                                                                                                                                    & \multicolumn{4}{c|}{\textbf{Query2Bid-10M}}                                                                                                                                                                   &                                                                                  &                                                                                                  \\ \cline{3-14}
                                   &  & \multicolumn{2}{c}{Exact}                                                                            & \multicolumn{2}{c|}{Phrase}                                                                           & \multicolumn{2}{c}{Exact}                                                                            & \multicolumn{2}{c|}{Phrase}                                                                           & \multicolumn{2}{c}{Exact}                                                                            & \multicolumn{2}{c|}{Phrase}                                                                           &                                                                                  &                                                                                                  \\ \cline{3-14}
     & \multirow{-3}{*}{\textbf{Model}} & \multicolumn{1}{l}{{Q@0.5}} & \multicolumn{1}{l}{{Q@0.7}} & \multicolumn{1}{l}{{Q@0.5}} & \multicolumn{1}{l|}{{Q@0.7}} & \multicolumn{1}{l}{{Q@0.5}} & \multicolumn{1}{l}{{Q@0.7}} & \multicolumn{1}{l}{{Q@0.5}} & \multicolumn{1}{l|}{{Q@0.7}} & \multicolumn{1}{l}{{Q@0.5}} & \multicolumn{1}{l}{{Q@0.7}} & \multicolumn{1}{l}{{Q@0.5}} & \multicolumn{1}{l|}{{Q@0.7}} & \multirow{-3}{*}{\begin{tabular}[c]{@{}c@{}}GPU \\ Latency\\  (ms)\end{tabular}} & \multirow{-3}{*}{\begin{tabular}[c]{@{}c@{}}Daily GPU\\ Cost for\\ 10k QPS\end{tabular}} \\ \Cline{2pt}{2-16}
    \parbox[t]{2mm}{\multirow{7}{*}{\rotatebox[origin=c]{90}{\textcolor{blue(ncs)}{\textit{Generative (NAR/AR)}}}}} & Transformer-base \cite{transformers}              & \multicolumn{1}{c}{1.56}                         & \multicolumn{1}{c}{0.76}                         & \multicolumn{1}{c}{5.93}                         & 4.18                                              & \multicolumn{1}{c}{4.13}                         & \multicolumn{1}{c}{2.05}                         & \multicolumn{1}{c}{16.24}                        & 10.91                                             & \multicolumn{1}{c}{7.10}                         & \multicolumn{1}{c}{3.56}                         & \multicolumn{1}{c}{23.42}                        & 18.95                                             & 108.8                                                                            & \$16,472                                                                                         \\ 
    & mT5-small \cite{mt5}                       & \multicolumn{1}{c}{1.66}                         & \multicolumn{1}{c}{0.82}                         & \multicolumn{1}{c}{6.34}                         & 4.48                                              & \multicolumn{1}{c}{4.49}                         & \multicolumn{1}{c}{2.22}                         & \multicolumn{1}{c}{15.51}                        & 11.89                                             & \multicolumn{1}{c}{7.64}                         & \multicolumn{1}{c}{3.70}                         & \multicolumn{1}{c}{24.94}                        & 20.24                                             & 168.94                                                                           & \$22,549                                                                                        \\ 
    & CLOVERv1 \cite{cloverv1}                        & \multicolumn{1}{c}{1.86}                         & \multicolumn{1}{c}{0.91}                         & \multicolumn{1}{c}{7.11}                         & 5.03                                              & \multicolumn{1}{c}{5.35}                         & \multicolumn{1}{c}{2.66}                         & \multicolumn{1}{c}{18.53}                        & 16.19                                             & \multicolumn{1}{c}{8.18}                         & \multicolumn{1}{c}{4.18}                         & \multicolumn{1}{c}{26.84}                        & 21.92                                             & 255.34                                                                           & \$38,658                                                                                         \\ 
    & Deep-Shallow \cite{deepshallow}                    & \multicolumn{1}{c}{1.33}                         & \multicolumn{1}{c}{0.64}                         & \multicolumn{1}{c}{5.21}                         & 3.67                                              & \multicolumn{1}{c}{3.76}                         & \multicolumn{1}{c}{1.87}                         & \multicolumn{1}{c}{13.11}                        & 9.97                                              & \multicolumn{1}{c}{6.13}                         & \multicolumn{1}{c}{3.01}                         & \multicolumn{1}{c}{20.82}                        & 16.85                                             & 75.09                                                                            & \$11,368                                                                                         \\ \cline{2-16}
     & CLOVERv2 (Beam 50)               & \multicolumn{1}{c}{1.54}                         & \multicolumn{1}{c}{0.71}                         & \multicolumn{1}{c}{6.57}                         & 4.32                                              & \multicolumn{1}{c}{3.76}                         & \multicolumn{1}{c}{1.75}                         & \multicolumn{1}{c}{15.30}                        & 10.61                                             & \multicolumn{1}{c}{5.91}                         & \multicolumn{1}{c}{2.87}                         & \multicolumn{1}{c}{22.65}                        & 15.49                                             & 5.64                                                                             & \$853                                                                                           \\ 
    & CLOVERv2 (Beam 100)                        & \multicolumn{1}{c}{2.49}                         & \multicolumn{1}{c}{1.19}                         & \multicolumn{1}{c}{9.81}                         & 6.64                                              & \multicolumn{1}{c}{5.86}                         & \multicolumn{1}{c}{2.74}                         & \multicolumn{1}{c}{23.91}                        & 16.60                                             & \multicolumn{1}{c}{8.18}                         & \multicolumn{1}{c}{3.90}                         & \multicolumn{1}{c}{31.94}                        & 22.65                                             & 5.64                                                                             & \$853                                                                                           \\ 
    & CLOVER-Unity (NLG)               & \multicolumn{1}{c}{2.97}                         & \multicolumn{1}{c}{1.45}                         & \multicolumn{1}{c}{11.71}                        & 7.78                                              & \multicolumn{1}{c}{6.95}                         & \multicolumn{1}{c}{3.20}                         & \multicolumn{1}{c}{28.41}                        & 19.61                                             & \multicolumn{1}{c}{9.82}                         & \multicolumn{1}{c}{4.68}                         & \multicolumn{1}{c}{38.32}                        & 27.19                                             & 5.64                                                                             & \$853                                                                                           \\ \cline{2-16}
    \parbox[t]{2mm}{\multirow{3}{*}{\rotatebox[origin=c]{90}{\textcolor{blue(ncs)}{\textit{Dense}}}}} & SimCSE \cite{simcse}                          & \multicolumn{1}{c}{3.34}                         & \multicolumn{1}{c}{1.58}                         & \multicolumn{1}{c}{10.15}                        & 6.97                                              & \multicolumn{1}{c}{9.16}                         & \multicolumn{1}{c}{4.40}                         & \multicolumn{1}{c}{26.23}                        & 18.99                                             & \multicolumn{1}{c}{16.49}                        & \multicolumn{1}{c}{7.02}                         & \multicolumn{1}{c}{41.87}                        & 30.97                                             & 4.78                                                                             & \$723                                                                                           \\ 
    & NGAME \cite{NGAME}                            & \multicolumn{1}{c}{3.45}                         & \multicolumn{1}{c}{1.63}                         & \multicolumn{1}{c}{10.47}                        & 7.19                                              & \multicolumn{1}{c}{9.45}                         & \multicolumn{1}{c}{4.55}                         & \multicolumn{1}{c}{27.05}                        & 19.58                                             & \multicolumn{1}{c}{16.94}                        & \multicolumn{1}{c}{7.24}                         & \multicolumn{1}{c}{43.17}                        & 31.93                                             & 4.78                                                                             & \$723                                                                                           \\ 
    & CLOVER-Unity (DR)                & \multicolumn{1}{c}{3.62}                         & \multicolumn{1}{c}{1.70}                         & \multicolumn{1}{c}{11.53}                        & 7.58                                              & \multicolumn{1}{c}{9.93}                         & \multicolumn{1}{c}{4.81}                         & \multicolumn{1}{c}{31.25}                        & 21.84                                             & \multicolumn{1}{c}{15.63}                        & \multicolumn{1}{c}{7.78}                         & \multicolumn{1}{c}{46.60}                        & 33.92                                             & 4.78                                                                             & \$723                                                                                           \\ \cline{2-16}
     & NGAME + CLOVERv2                 & \multicolumn{1}{c}{4.11}                         & \multicolumn{1}{c}{1.85}                         & \multicolumn{1}{c}{15.54}                        & 10.03                                             & \multicolumn{1}{c}{11.59}                        & \multicolumn{1}{c}{5.25}                         & \multicolumn{1}{c}{41.73}                        & 28.47                                             & \multicolumn{1}{c}{17.91}                        & \multicolumn{1}{c}{8.29}                         & \multicolumn{1}{c}{61.62}                        & 43.43                                             & 5.64                                                                             & \$1,577                                                                                          \\ 
  \rowcolor{Gray}  & \textbf{CLOVER-Unity}            & \multicolumn{1}{c}{\textbf{4.43}}                & \multicolumn{1}{c}{\textbf{2.07}}                & \multicolumn{1}{c}{\textbf{17.08}}               & \textbf{10.90}                                    & \multicolumn{1}{c}{\textbf{12.79}}               & \multicolumn{1}{c}{\textbf{5.74}}                & \multicolumn{1}{c}{\textbf{46.60}}               & \textbf{31.39}                                    & \multicolumn{1}{c}{\textbf{19.60}}               & \multicolumn{1}{c}{\textbf{9.25}}                & \multicolumn{1}{c}{\textbf{66.93}}               & \textbf{46.40}                                    & \textbf{5.64}                                                                    & \$853                                                                                           \\ \Cline{2pt}{2-16}
    \end{tabular}}
    \caption{Comparison of performance and cost metrics for NLG, DR, and Unity retrieval models on sponsored search datasets}
    \label{table:offline_results_sponsored_search}
    \endgroup
    \end{table*}
Table \ref{table:offline_results_sponsored_search} shows the results on the three sponsored search datasets. We observe that AR NLG models incur high latency and computation costs. For example, DeepShallow \cite{deepshallow}, which uses an efficient architecture with a 12-layer encoder and a 1-layer decoder, is still 15 times slower and more expensive than DR models. On the other hand, CLOVERv2, a NAR model, retrieves more good keywords than AR models by leveraging larger beam sizes while being much more efficient. Figure \ref{fig:latency_beam} shows that the mean inference latency of CLOVERv2 is almost constant (around 10-12 ms) for any beam size, in contrast to AR models. Compared to CLOVERv2, DR models such as NGAME perform better or on par in most datasets. However, CLOVERv2 and NGAME retrieve different good keywords. About 40\% of the good keywords retrieved by CLOVERv2 were not retrieved by NGAME and vice versa. Therefore, the ensemble of NGAME and CLOVERv2 (union of individual keywords) outperforms the best of either method by an average of 17.5\% and 44.3\% in exact and phrase good keyword densities.

We further analyzed the differences in keywords retrieved by NLG and DR approaches and found two key observations. First, NLG models more effectively captured word-level relationships between the query and keywords, such as synonymy and hypernymy. We used the WordNet lexical database \cite{wordnet} to compute the average number of words in the retrieved keywords that were synonyms or hypernyms of query words. Figure \ref{subsec:generative_vs_dense} (left) shows that keywords from CLOVERv2 contained 36.7\% and 34.2\% more synonyms and hypernyms than those from NGAME, respectively. This advantage likely resulted from NLG models having an explicit token-level loss, which DR models lack. Furthermore, as shown in Figure \ref{subsec:generative_vs_dense} (right), CLOVERv2 tends to produce significantly shorter keywords than NGAME. This discrepancy stemmed from its non-autoregressive nature, which requires correctly predicting all $n$ keyword tokens in parallel, unlike DR methods that don't rely on keyword length. 
\begin{figure}
     \centering
     \begin{subfigure}[b]{0.49\linewidth}
         \centering
         \includegraphics[width=\textwidth]{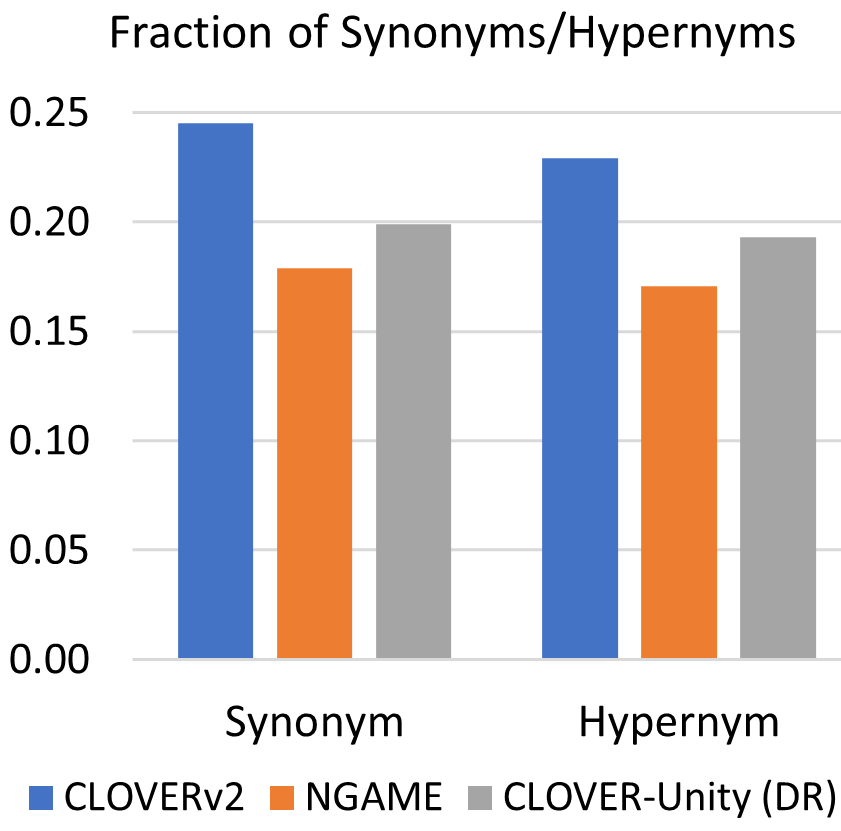}
         \label{fig:gen_dense_analysis_syn}
     \end{subfigure}
     \hfill
     \begin{subfigure}[b]{0.48\linewidth}
         \centering
         \includegraphics[width=\textwidth]{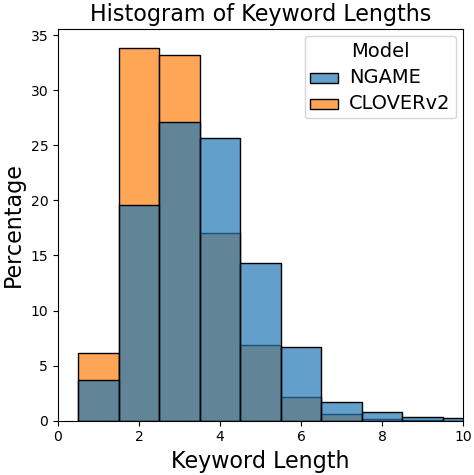}
         \label{fig:gen_dense_analysis_len}
     \end{subfigure}
    \caption{Synonymy and hypernymy relations between query and keyword (left) and keyword length distribution (right)}
    \label{fig:gen_dense_analysis}
\end{figure}
\subsection{Results of Unified Retrieval}
\label{subsec:unified_retrieval}
As shown in table \ref{table:offline_results_sponsored_search} (rows 7 and 10), the NLG and DR components of CLOVER-Unity outperform the best-performing baselines, CLOVERv2 and NGAME, by an average of 19.2\% and 7.4\%, respectively. Further, CLOVER-Unity, with both its components (last row), improves upon the ensemble of CLOVERv2 and NGAME by 9.8\% on average while reducing the GPU compute costs by almost 45\%. These results highlight the effectiveness of our proposed approach in unifying the benefits of generative and dense retrieval methods in a single model. Interestingly, we observe that keywords retrieved by the DR component of CLOVER-Unity contain 11.2\% and 13.1\% more synonyms and hypernyms than NGAME (figure \ref{fig:gen_dense_analysis}). These findings support our hypothesis that models trained with token-level loss are better at identifying such word-level relationships.  
\subsection{Human Evaluations}
\label{subsec:human_eval}
We conducted human evaluations on CLOVER-Unity (DR) and NGAME using 500 English, French, and German queries. We sampled one keyword per query after filtering them using a lightweight relevance model. We asked multiple annotators to rate each query-keyword pair as “good” or “bad” based on the exact and phrase match quality. We report the percentage of query keyword pairs labeled "good" for the different models and match types in Table \ref{tab:human_eval}. CLOVER-Unity (DR) outperformed NGAME across all languages and match types, with a relative improvement of 8.94\% and 12.58\% in exact and phrase match, respectively.
\subsection{Public Benchmarks}
\label{public_benchmarks}
We present the results on the public short-text benchmarks in table \ref{tab:results_public}. NGAME outperforms all other leading extreme classification or DR methods, such as ECLARE \cite{eclare}, GalaXC \cite{galaxc}, and DECAF \cite{decaf}. However, CLOVER-Unity (DR), which uses our proposed unified retrieval approach, surpasses the strong NGAME baseline on all metrics and datasets. It achieves an average improvement of 4.4\% and 3.2\% in PSP@5 and Recall@100, respectively. These results highlight the versatility of CLOVER-Unity and its applicability to tasks beyond query rewriting in sponsored search. 
\begin{table}[]
    \centering
    \begingroup
    \def\arraystretch{1.05}
    \resizebox{0.85\linewidth}{!}{
\begin{tabular}{l|cc|cc}
\hline
\multicolumn{1}{c|}{\multirow{2}{*}{Language}} & \multicolumn{2}{c|}{Exact Match}                             & \multicolumn{2}{c}{Phrase Match}                            \\ \cline{2-5} 
\multicolumn{1}{c|}{}                          & \multicolumn{1}{c}{NGAME}  & \multicolumn{1}{l|}{Unity} & \multicolumn{1}{c}{NGAME}  & \multicolumn{1}{l}{Unity} \\ \hline
English                                         & \multicolumn{1}{c}{32.8\%} & 37.4\%                         & \multicolumn{1}{c}{57.2\%} & 59.8\%                         \\ \hline
French                                          & \multicolumn{1}{c}{40.8\%} & 44.0\%                         & \multicolumn{1}{c}{40.8\%} & 45.0\%                         \\ \hline
German                                          & \multicolumn{1}{c}{36.0\%} & 38.0\%                         & \multicolumn{1}{c}{31.8\%} & 35.8\%                         \\ \hline
\end{tabular}}
\caption{Proportion of keywords from NGAME and CLOVER-Unity (DR) labeled as high-quality by human annotators}
\label{tab:human_eval}
\endgroup
\end{table}

\section{Online Experiments}
\begin{figure}
    \centering
    \includegraphics[width=0.45\textwidth]{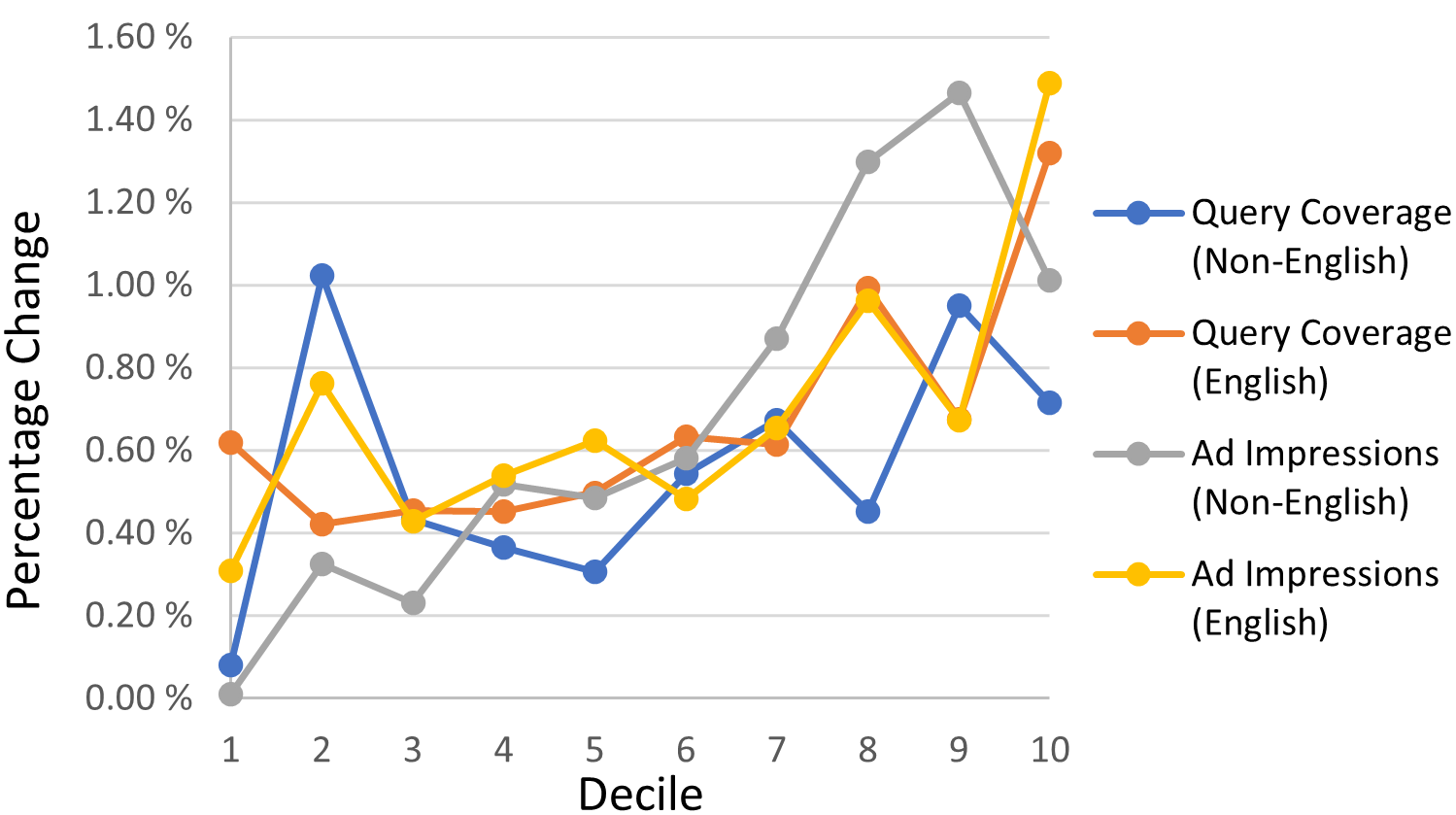}
    \caption{Change in query coverage and Ad impression for different deciles in online A/B tests on Microsoft Bing}
    \label{fig:online_decile_analysis}
\end{figure}
We conducted extensive online experiments on the live traffic of Microsoft Bing across 140+ countries. We deployed CLOVER-Unity with its DR and NLG components and compared them with an ensemble of leading (proprietary) DR, NLG, large language models, extreme classification, and graph-based techniques in production (control). The data was collected through A/B testing over a substantial proportion of traffic and time.	We used three primary metrics:
\begin{itemize}
    \item \textbf{Clicks} denote the total user  clicks on ads. This measures user engagement and directly contributes to revenue.
    \item \textbf{Quick Back Rate} (QBR) denotes the percentage of ad clicks that resulted in a quick back with low dwell time. 
    \item \textbf{Ad Defect} denotes the percentage of bad ads, as measured by offline relevance models.     
\end{itemize}
\begin{table}[]
    \centering
    \begingroup
    \def\arraystretch{1.19}
    \resizebox{0.96\linewidth}{!}{
    \begin{tabular}{lccccc}
    \Xhline{2\arrayrulewidth}
    \multicolumn{1}{l|}{Model}             & \multicolumn{1}{c|}{PSP@5} & \multicolumn{1}{c|}{P@5}   & \multicolumn{1}{c|}{PSN@5} & \multicolumn{1}{c|}{N@5}   & R@100 \\ \hline
    \multicolumn{6}{c}{WikiSeeAlsoTitles}                                                                                                                             \\ \hline
    \multicolumn{1}{l|}{DECAF \cite{decaf}}             & \multicolumn{1}{c|}{21.01} & \multicolumn{1}{c|}{12.86} & \multicolumn{1}{c|}{20.75} & \multicolumn{1}{c|}{25.95} & -     \\ 
    \multicolumn{1}{l|}{ECLARE \cite{eclare}}            & \multicolumn{1}{c|}{26.27} & \multicolumn{1}{c|}{15.05} & \multicolumn{1}{c|}{26.03} & \multicolumn{1}{c|}{30.20}  & -     \\ 
    \multicolumn{1}{l|}{GalaXC \cite{galaxc}}            & \multicolumn{1}{c|}{24.47} & \multicolumn{1}{c|}{14.30}  & \multicolumn{1}{c|}{23.16} & \multicolumn{1}{c|}{27.6}  & -     \\ 
    \multicolumn{1}{l|}{NGAME M1 \cite{NGAME}}          & \multicolumn{1}{c|}{30.61} & \multicolumn{1}{c|}{16.04} & \multicolumn{1}{c|}{30.63} & \multicolumn{1}{c|}{32.17} & 54.08 \\ 
    \rowcolor{Gray} \multicolumn{1}{l|}{\textbf{CLOVER-Unity (DR)}} & \multicolumn{1}{c|}{\textbf{32.38}} & \multicolumn{1}{c|}{\textbf{16.57}} & \multicolumn{1}{c|}{\textbf{32.10}}  & \multicolumn{1}{c|}{\textbf{32.97}} & \textbf{55.84} \\ \hline
    \multicolumn{6}{c}{AmazonTitles}                                                                                                                                  \\ \hline
    \multicolumn{1}{l|}{DECAF \cite{decaf}}             & \multicolumn{1}{c|}{41.42} & \multicolumn{1}{c|}{18.65} & \multicolumn{1}{c|}{37.13} & \multicolumn{1}{c|}{41.46} & -     \\ 
    \multicolumn{1}{l|}{ECLARE \cite{eclare}}            & \multicolumn{1}{c|}{44.7}  & \multicolumn{1}{c|}{19.88} & \multicolumn{1}{c|}{40.21} & \multicolumn{1}{c|}{44.16} & -     \\ 
    \multicolumn{1}{l|}{GalaXC \cite{galaxc}}            & \multicolumn{1}{c|}{43.95} & \multicolumn{1}{c|}{19.49} & \multicolumn{1}{c|}{39.37} & \multicolumn{1}{c|}{43.06} & -     \\ 
    \multicolumn{1}{l|}{NGAME M1 \cite{NGAME}}          & \multicolumn{1}{c|}{47.63} & \multicolumn{1}{c|}{20.36} & \multicolumn{1}{c|}{43.83} & \multicolumn{1}{c|}{45.60}  & 67.22 \\ 
   \rowcolor{Gray} \multicolumn{1}{l|}{\textbf{CLOVER-Unity (DR)}} & \multicolumn{1}{c|}{\textbf{49.03}} & \multicolumn{1}{c|}{\textbf{20.9}}  & \multicolumn{1}{c|}{\textbf{44.706}} & \multicolumn{1}{c|}{\textbf{46.38}} & \textbf{69.34} \\ \Xhline{2\arrayrulewidth}
    \end{tabular}}
    \caption{Comparison of DR component of CLOVER-Unity and extreme classification methods on public benchmarks}
    \label{tab:results_public}
    \endgroup
    \end{table}
\begin{table}[]
\centering
\begingroup
\def\arraystretch{1}
\resizebox{0.95\linewidth}{!}{
\begin{tabular}{l|c|c|c|c}
\hline
{Language} & {$\Delta$ Revenue} & {$\Delta$ Clicks} & {$\Delta$ QBR} & {$\Delta$ Defect} \\ \hline
English                   & \textbf{\textcolor{ForestGreen}{1.12\%}}           & \textbf{\textcolor{ForestGreen}{0.86\%}}            & \textcolor{gray}{-0.19\%}  & \textcolor{gray}{0.12\%}      \\ \hline
Non-english               & \textbf{\textcolor{ForestGreen}{1.43\%}}           & \textbf{\textcolor{ForestGreen}{0.93\%}}          & \textcolor{gray}{0.27\%} & \textcolor{gray}{0.16\%}      \\ \hline
\end{tabular}}
\caption{Results of online A/B tests on Microsoft Bing, with gray color indicating non-significant results (p-value > 0.01)}
\label{tab:online_results}
\endgroup
\end{table}

Our objective is to maximize clicks without increasing QBR and ad defect rate. Table \ref{tab:online_results} summarizes the results of the stated metric for English and non-English queries. CLOVER-Unity increased overall clicks (0.93\% in non-English and 0.86\% in English segments) without any statistically significant change in QBR. This indicates that the revenue increase, 1.43\% in non-English and 1.12\% in English segments, resulted from healthy clicks on sponsored content that users found relevant. We also don't see any statistical change in the ad defect, indicating that the ads shown were relevant. To further analyze these gains, we compared query coverage and ad impression of our model with control models. We grouped queries into buckets, also called deciles in sponsored search; each decile contains queries with a similar frequency. For instance, decile 1 contains highly frequent queries with a small total count (usually less than 100). In contrast, decile 10 contains rare queries with a large total count (usually in the millions) because of the long tail distribution of search queries. We measured \textbf{query coverage} for a decile as the proportion of queries in the decile for which any sponsored content was displayed. \textbf{Ad impression} is the total number of ads shown for queries in a decile. Figure \ref{fig:online_decile_analysis} shows that our model had higher query coverage for both English and non-English queries, indicating that we displayed relevant sponsored content for search queries not served before. We obtained query coverage increases on all deciles with better gains on tail queries, which are usually longer, more ambiguous, and harder to cater to. For instance, as shown in Table \ref{tab:Example}, for the tail query \textit{"currys 912 black"}, existing DR and NLG methods failed to understand the user’s intent. But CLOVER-Unity retrieved high-quality keywords matching the underlying user intent. We saw a similar pattern for ad impressions, indicating that we served more sponsored content relevant to users for all query deciles, with more pronounced gains on tail queries.
\section{Practical Recommendation}
\label{sec:practical_lessons}
We summarize the key insights from our study and provide practical recommendations and optimization tips. 
\begin{enumerate}
    \item Use a unified encoder for both NLG and DR instead of separate models. However, apply different quality thresholds (downstream) to filter keywords from the NLG and DR outputs, as they may vary in quality.   
    \item Cluster keywords using DR embeddings to group similar ones with minor syntactic variations. This reduced our DiskANN index size and, consequently, the CPU RAM and SSD memory requirement by around 3x. 
    \item Use early pruning strategies in trie beam search to improve quality and latency. For example, we remove a partially generated keyword from the beam if the log probability of any token is below a threshold. This reduced the 99.9th percentile latency by 3x in our study. 
    \item Offload trie-beam search to a separate CPU server. This ensures that GPUs are only used for the model forward pass, reducing the GPU idle time and improving GPU throughput. 
    \item In our experiments, Onnxruntime achieved the lowest inference latency compared to other engines such as Torch JIT, Tensorflow, FasterTransformers, and TVM. Nvidia Triton Inference server reduced overall latency by 2ms. Further, A100s with Multi-Instance GPU (MIG) partitioning lowered GPU cost by 2.2x over T4 GPUs.
\end{enumerate}
\section{Conclusion}
In this work, we addressed the problem of retrieving high-quality exact and phrase bid keywords for user search queries. We contrasted two different approaches to this problem: dense and generative retrieval. We showed that the two methods retrieved different sets of high-quality keywords for the same query, suggesting that their performance was complimentary. To combine their advantages, we proposed CLOVER-Unity, which optimizes a shared encoder for both DR and NLG. We evaluated our approach through extensive offline and online experiments. For instance, we showed that CLOVER-Unity outperformed the ensemble of two separate NLG and DR models with half the GPU compute. We also shared our key insights and optimization tips from our study.


\newpage
\bibliographystyle{ACM-Reference-Format}
\bibliography{references}


\begin{thebibliography}{31}


\ifx \showCODEN    \undefined \def \showCODEN     #1{\unskip}     \fi
\ifx \showDOI      \undefined \def \showDOI       #1{#1}\fi
\ifx \showISBNx    \undefined \def \showISBNx     #1{\unskip}     \fi
\ifx \showISBNxiii \undefined \def \showISBNxiii  #1{\unskip}     \fi
\ifx \showISSN     \undefined \def \showISSN      #1{\unskip}     \fi
\ifx \showLCCN     \undefined \def \showLCCN      #1{\unskip}     \fi
\ifx \shownote     \undefined \def \shownote      #1{#1}          \fi
\ifx \showarticletitle \undefined \def \showarticletitle #1{#1}   \fi
\ifx \showURL      \undefined \def \showURL       {\relax}        \fi
\providecommand\bibfield[2]{#2}
\providecommand\bibinfo[2]{#2}
\providecommand\natexlab[1]{#1}
\providecommand\showeprint[2][]{arXiv:#2}

\bibitem[\protect\citeauthoryear{Bai, Ordentlich, Zhang, Feng, Ratnaparkhi,
  Somvanshi, and Tjahjadi}{Bai et~al\mbox{.}}{2018}]%
        {Bai2018ScalableQN}
\bibfield{author}{\bibinfo{person}{Xiao Bai}, \bibinfo{person}{E. Ordentlich},
  \bibinfo{person}{Yuanyuan Zhang}, \bibinfo{person}{Andy Feng},
  \bibinfo{person}{A. Ratnaparkhi}, \bibinfo{person}{Reena Somvanshi}, {and}
  \bibinfo{person}{Aldi Tjahjadi}.} \bibinfo{year}{2018}\natexlab{}.
\newblock \showarticletitle{Scalable Query N-Gram Embedding for Improving
  Matching and Relevance in Sponsored Search}.
\newblock \bibinfo{journal}{\emph{Proceedings of the 24th ACM SIGKDD
  International Conference on Knowledge Discovery \& Data Mining}}
  (\bibinfo{year}{2018}).
\newblock


\bibitem[\protect\citeauthoryear{Bhatia, Dahiya, Jain, Kar, Mittal, Prabhu, and
  Varma}{Bhatia et~al\mbox{.}}{2016}]%
        {manik_xc_repo}
\bibfield{author}{\bibinfo{person}{K. Bhatia}, \bibinfo{person}{K. Dahiya},
  \bibinfo{person}{H. Jain}, \bibinfo{person}{P. Kar}, \bibinfo{person}{A.
  Mittal}, \bibinfo{person}{Y. Prabhu}, {and} \bibinfo{person}{M. Varma}.}
  \bibinfo{year}{2016}\natexlab{}.
\newblock \bibinfo{title}{The extreme classification repository: Multi-label
  datasets and code}.
\newblock
\newblock
\urldef\tempurl%
\url{http://manikvarma.org/downloads/XC/XMLRepository.html}
\showURL{%
\tempurl}


\bibitem[\protect\citeauthoryear{Broder, Ciccolo, Fontoura, Gabrilovich,
  Josifovski, and Riedel}{Broder et~al\mbox{.}}{2008}]%
        {Broder2008SearchAU}
\bibfield{author}{\bibinfo{person}{A. Broder}, \bibinfo{person}{P. Ciccolo},
  \bibinfo{person}{Marcus Fontoura}, \bibinfo{person}{Evgeniy Gabrilovich},
  \bibinfo{person}{V. Josifovski}, {and} \bibinfo{person}{L. Riedel}.}
  \bibinfo{year}{2008}\natexlab{}.
\newblock \showarticletitle{Search advertising using web relevance feedback}.
  In \bibinfo{booktitle}{\emph{CIKM '08}}.
\newblock


\bibitem[\protect\citeauthoryear{Broder, Fontoura, Josifovski, and
  Riedel}{Broder et~al\mbox{.}}{2007}]%
        {Broder2007ASA}
\bibfield{author}{\bibinfo{person}{A. Broder}, \bibinfo{person}{Marcus
  Fontoura}, \bibinfo{person}{V. Josifovski}, {and} \bibinfo{person}{L.
  Riedel}.} \bibinfo{year}{2007}\natexlab{}.
\newblock \showarticletitle{A semantic approach to contextual advertising}. In
  \bibinfo{booktitle}{\emph{SIGIR}}.
\newblock


\bibitem[\protect\citeauthoryear{Conneau, Khandelwal, Goyal, Chaudhary, Wenzek,
  Guzm{\'a}n, Grave, Ott, Zettlemoyer, and Stoyanov}{Conneau
  et~al\mbox{.}}{2020}]%
        {xlmr}
\bibfield{author}{\bibinfo{person}{Alexis Conneau}, \bibinfo{person}{Kartikay
  Khandelwal}, \bibinfo{person}{Naman Goyal}, \bibinfo{person}{Vishrav
  Chaudhary}, \bibinfo{person}{Guillaume Wenzek}, \bibinfo{person}{Francisco
  Guzm{\'a}n}, \bibinfo{person}{E. Grave}, \bibinfo{person}{Myle Ott},
  \bibinfo{person}{Luke Zettlemoyer}, {and} \bibinfo{person}{Veselin
  Stoyanov}.} \bibinfo{year}{2020}\natexlab{}.
\newblock \showarticletitle{Unsupervised Cross-lingual Representation Learning
  at Scale}. In \bibinfo{booktitle}{\emph{ACL}}.
\newblock


\bibitem[\protect\citeauthoryear{Dahiya, Agarwal, Saini, Gururaj, Jiao, Singh,
  Agarwal, Kar, and Varma}{Dahiya et~al\mbox{.}}{2021}]%
        {SiameseXML}
\bibfield{author}{\bibinfo{person}{Kunal Dahiya}, \bibinfo{person}{Ananye
  Agarwal}, \bibinfo{person}{Deepak Saini}, \bibinfo{person}{K Gururaj},
  \bibinfo{person}{Jian Jiao}, \bibinfo{person}{Amit Singh},
  \bibinfo{person}{Sumeet Agarwal}, \bibinfo{person}{Purushottam Kar}, {and}
  \bibinfo{person}{Manik Varma}.} \bibinfo{year}{2021}\natexlab{}.
\newblock \showarticletitle{SiameseXML: Siamese Networks meet Extreme
  Classifiers with 100M Labels}. In \bibinfo{booktitle}{\emph{ICML}}.
\newblock


\bibitem[\protect\citeauthoryear{Dahiya, Gupta, Saini, Soni, Wang, Dave, Jiao,
  Gururaj, Dey, Singh, Hada, Jain, Paliwal, Mittal, Mehta, Ramjee, Agarwal,
  Kar, and Varma}{Dahiya et~al\mbox{.}}{2022}]%
        {NGAME}
\bibfield{author}{\bibinfo{person}{Kunal Dahiya}, \bibinfo{person}{Nilesh
  Gupta}, \bibinfo{person}{Deepak Saini}, \bibinfo{person}{Akshay Soni},
  \bibinfo{person}{Yajun Wang}, \bibinfo{person}{Kushal Dave},
  \bibinfo{person}{Jian Jiao}, \bibinfo{person}{K Gururaj},
  \bibinfo{person}{Prasenjit Dey}, \bibinfo{person}{Amit Singh},
  \bibinfo{person}{Deepesh~V. Hada}, \bibinfo{person}{Vidit Jain},
  \bibinfo{person}{Bhawna Paliwal}, \bibinfo{person}{Ansh Mittal},
  \bibinfo{person}{Sonu Mehta}, \bibinfo{person}{Ramachandran Ramjee},
  \bibinfo{person}{Sumeet Agarwal}, \bibinfo{person}{Purushottam Kar}, {and}
  \bibinfo{person}{Manik Varma}.} \bibinfo{year}{2022}\natexlab{}.
\newblock \showarticletitle{NGAME: Negative Mining-aware Mini-batching for
  Extreme Classification}.
\newblock \bibinfo{journal}{\emph{ArXiv}}  \bibinfo{volume}{abs/2207.04452}
  (\bibinfo{year}{2022}).
\newblock


\bibitem[\protect\citeauthoryear{Devlin, Chang, Lee, and Toutanova}{Devlin
  et~al\mbox{.}}{2019}]%
        {bert}
\bibfield{author}{\bibinfo{person}{Jacob Devlin}, \bibinfo{person}{Ming-Wei
  Chang}, \bibinfo{person}{Kenton Lee}, {and} \bibinfo{person}{Kristina
  Toutanova}.} \bibinfo{year}{2019}\natexlab{}.
\newblock \showarticletitle{BERT: Pre-training of Deep Bidirectional
  Transformers for Language Understanding}.
\newblock \bibinfo{journal}{\emph{ArXiv}}  \bibinfo{volume}{abs/1810.04805}
  (\bibinfo{year}{2019}).
\newblock


\bibitem[\protect\citeauthoryear{Gao, Yao, and Chen}{Gao et~al\mbox{.}}{2021}]%
        {simcse}
\bibfield{author}{\bibinfo{person}{Tianyu Gao}, \bibinfo{person}{Xingcheng
  Yao}, {and} \bibinfo{person}{Danqi Chen}.} \bibinfo{year}{2021}\natexlab{}.
\newblock \showarticletitle{SimCSE: Simple Contrastive Learning of Sentence
  Embeddings}.
\newblock \bibinfo{journal}{\emph{ArXiv}}  \bibinfo{volume}{abs/2104.08821}
  (\bibinfo{year}{2021}).
\newblock


\bibitem[\protect\citeauthoryear{Gu, Bradbury, Xiong, Li, and Socher}{Gu
  et~al\mbox{.}}{2017}]%
        {NAR}
\bibfield{author}{\bibinfo{person}{Jiatao Gu}, \bibinfo{person}{James
  Bradbury}, \bibinfo{person}{Caiming Xiong}, \bibinfo{person}{Victor O.~K.
  Li}, {and} \bibinfo{person}{Richard Socher}.}
  \bibinfo{year}{2017}\natexlab{}.
\newblock \showarticletitle{Non-Autoregressive Neural Machine Translation}.
\newblock \bibinfo{journal}{\emph{ArXiv}}  \bibinfo{volume}{abs/1711.02281}
  (\bibinfo{year}{2017}).
\newblock


\bibitem[\protect\citeauthoryear{Hofst{\"a}tter, Lin, Yang, Lin, and
  Hanbury}{Hofst{\"a}tter et~al\mbox{.}}{2021}]%
        {TAS}
\bibfield{author}{\bibinfo{person}{Sebastian Hofst{\"a}tter},
  \bibinfo{person}{Sheng-Chieh Lin}, \bibinfo{person}{Jheng-Hong Yang},
  \bibinfo{person}{Jimmy~J. Lin}, {and} \bibinfo{person}{Allan Hanbury}.}
  \bibinfo{year}{2021}\natexlab{}.
\newblock \showarticletitle{Efficiently Teaching an Effective Dense Retriever
  with Balanced Topic Aware Sampling}.
\newblock \bibinfo{journal}{\emph{Proceedings of the 44th International ACM
  SIGIR Conference on Research and Development in Information Retrieval}}
  (\bibinfo{year}{2021}).
\newblock


\bibitem[\protect\citeauthoryear{Jain, Prabhu, and Varma}{Jain
  et~al\mbox{.}}{2016}]%
        {xc_metrics}
\bibfield{author}{\bibinfo{person}{Himanshu Jain}, \bibinfo{person}{Yashoteja
  Prabhu}, {and} \bibinfo{person}{Manik Varma}.}
  \bibinfo{year}{2016}\natexlab{}.
\newblock \showarticletitle{Extreme Multi-label Loss Functions for
  Recommendation, Tagging, Ranking \& Other Missing Label Applications}.
\newblock \bibinfo{journal}{\emph{Proceedings of the 22nd ACM SIGKDD
  International Conference on Knowledge Discovery and Data Mining}}
  (\bibinfo{year}{2016}).
\newblock


\bibitem[\protect\citeauthoryear{Jayaram~Subramanya, Devvrit, Simhadri,
  Krishnawamy, and Kadekodi}{Jayaram~Subramanya et~al\mbox{.}}{2019}]%
        {DiskANN}
\bibfield{author}{\bibinfo{person}{Suhas Jayaram~Subramanya},
  \bibinfo{person}{Fnu Devvrit}, \bibinfo{person}{Harsha~Vardhan Simhadri},
  \bibinfo{person}{Ravishankar Krishnawamy}, {and} \bibinfo{person}{Rohan
  Kadekodi}.} \bibinfo{year}{2019}\natexlab{}.
\newblock \showarticletitle{DiskANN: Fast Accurate Billion-point Nearest
  Neighbor Search on a Single Node}. In \bibinfo{booktitle}{\emph{Advances in
  Neural Information Processing Systems}},
  \bibfield{editor}{\bibinfo{person}{H.~Wallach},
  \bibinfo{person}{H.~Larochelle}, \bibinfo{person}{A.~Beygelzimer},
  \bibinfo{person}{F.~d\textquotesingle Alch\'{e}-Buc},
  \bibinfo{person}{E.~Fox}, {and} \bibinfo{person}{R.~Garnett}} (Eds.),
  Vol.~\bibinfo{volume}{32}. \bibinfo{publisher}{Curran Associates, Inc.}
\newblock
\urldef\tempurl%
\url{https://proceedings.neurips.cc/paper/2019/file/09853c7fb1d3f8ee67a61b6bf4a7f8e6-Paper.pdf}
\showURL{%
\tempurl}


\bibitem[\protect\citeauthoryear{Kasai, Pappas, Peng, Cross, and Smith}{Kasai
  et~al\mbox{.}}{2020}]%
        {deepshallow}
\bibfield{author}{\bibinfo{person}{Jungo Kasai}, \bibinfo{person}{Nikolaos
  Pappas}, \bibinfo{person}{Hao Peng}, \bibinfo{person}{James Cross}, {and}
  \bibinfo{person}{Noah~A. Smith}.} \bibinfo{year}{2020}\natexlab{}.
\newblock \showarticletitle{Deep Encoder, Shallow Decoder: Reevaluating
  Non-autoregressive Machine Translation}. In
  \bibinfo{booktitle}{\emph{International Conference on Learning
  Representations}}.
\newblock


\bibitem[\protect\citeauthoryear{Miller}{Miller}{1995}]%
        {wordnet}
\bibfield{author}{\bibinfo{person}{George~A. Miller}.}
  \bibinfo{year}{1995}\natexlab{}.
\newblock \showarticletitle{WordNet: A Lexical Database for English}.
\newblock \bibinfo{journal}{\emph{Commun. ACM}}  \bibinfo{volume}{38}
  (\bibinfo{year}{1995}), \bibinfo{pages}{39--41}.
\newblock


\bibitem[\protect\citeauthoryear{Mittal, Dahiya, Agrawal, Saini, Agarwal, Kar,
  and Varma}{Mittal et~al\mbox{.}}{2021a}]%
        {decaf}
\bibfield{author}{\bibinfo{person}{Anshul Mittal}, \bibinfo{person}{Kunal
  Dahiya}, \bibinfo{person}{Sheshansh Agrawal}, \bibinfo{person}{Deepak Saini},
  \bibinfo{person}{Sumeet Agarwal}, \bibinfo{person}{Purushottam Kar}, {and}
  \bibinfo{person}{Manik Varma}.} \bibinfo{year}{2021}\natexlab{a}.
\newblock \showarticletitle{DECAF: Deep Extreme Classification with Label
  Features}.
\newblock \bibinfo{journal}{\emph{Proceedings of the 14th ACM International
  Conference on Web Search and Data Mining}} (\bibinfo{year}{2021}).
\newblock


\bibitem[\protect\citeauthoryear{Mittal, Sachdeva, Agrawal, Agarwal, Kar, and
  Varma}{Mittal et~al\mbox{.}}{2021b}]%
        {eclare}
\bibfield{author}{\bibinfo{person}{Anshul Mittal}, \bibinfo{person}{Noveen
  Sachdeva}, \bibinfo{person}{Sheshansh Agrawal}, \bibinfo{person}{Sumeet
  Agarwal}, \bibinfo{person}{Purushottam Kar}, {and} \bibinfo{person}{Manik
  Varma}.} \bibinfo{year}{2021}\natexlab{b}.
\newblock \showarticletitle{ECLARE: Extreme Classification with Label Graph
  Correlations}.
\newblock \bibinfo{journal}{\emph{Proceedings of the Web Conference 2021}}
  (\bibinfo{year}{2021}).
\newblock


\bibitem[\protect\citeauthoryear{Mohankumar, Begwani, and Singh}{Mohankumar
  et~al\mbox{.}}{2021}]%
        {cloverv1}
\bibfield{author}{\bibinfo{person}{Akash~Kumar Mohankumar},
  \bibinfo{person}{Nikit Begwani}, {and} \bibinfo{person}{Amit Singh}.}
  \bibinfo{year}{2021}\natexlab{}.
\newblock \showarticletitle{Diversity driven Query Rewriting in Search
  Advertising}.
\newblock \bibinfo{journal}{\emph{Proceedings of the 27th ACM SIGKDD Conference
  on Knowledge Discovery \& Data Mining}} (\bibinfo{year}{2021}).
\newblock


\bibitem[\protect\citeauthoryear{Papineni, Roukos, Ward, and Zhu}{Papineni
  et~al\mbox{.}}{2002}]%
        {bleu}
\bibfield{author}{\bibinfo{person}{Kishore Papineni}, \bibinfo{person}{Salim
  Roukos}, \bibinfo{person}{Todd Ward}, {and} \bibinfo{person}{Wei-Jing Zhu}.}
  \bibinfo{year}{2002}\natexlab{}.
\newblock \showarticletitle{{BLEU:} a Method for Automatic Evaluation of
  Machine Translation}. In \bibinfo{booktitle}{\emph{Proceedings of the 40th
  Annual Meeting of the Association for Computational Linguistics}}.
  \bibinfo{publisher}{Association for Computational Linguistics},
  \bibinfo{address}{Philadelphia, Pennsylvania, USA},
  \bibinfo{pages}{311--318}.
\newblock
\urldef\tempurl%
\url{https://doi.org/10.3115/1073083.1073135}
\showDOI{\tempurl}


\bibitem[\protect\citeauthoryear{Prabhu, Kag, Gopinath, Dahiya, Harsola,
  Agrawal, and Varma}{Prabhu et~al\mbox{.}}{2018a}]%
        {Prabhu2018ExtremeML}
\bibfield{author}{\bibinfo{person}{Yashoteja Prabhu}, \bibinfo{person}{Anil
  Kag}, \bibinfo{person}{Shilpa Gopinath}, \bibinfo{person}{Kunal Dahiya},
  \bibinfo{person}{Shrutendra Harsola}, \bibinfo{person}{Rahul Agrawal}, {and}
  \bibinfo{person}{Manik Varma}.} \bibinfo{year}{2018}\natexlab{a}.
\newblock \showarticletitle{Extreme Multi-label Learning with Label Features
  for Warm-start Tagging, Ranking \& Recommendation}.
\newblock \bibinfo{journal}{\emph{Proceedings of the Eleventh ACM International
  Conference on Web Search and Data Mining}} (\bibinfo{year}{2018}).
\newblock


\bibitem[\protect\citeauthoryear{Prabhu, Kag, Harsola, Agrawal, and
  Varma}{Prabhu et~al\mbox{.}}{2018b}]%
        {parabel}
\bibfield{author}{\bibinfo{person}{Yashoteja Prabhu}, \bibinfo{person}{Anil
  Kag}, \bibinfo{person}{Shrutendra Harsola}, \bibinfo{person}{Rahul Agrawal},
  {and} \bibinfo{person}{Manik Varma}.} \bibinfo{year}{2018}\natexlab{b}.
\newblock \showarticletitle{Parabel: Partitioned Label Trees for Extreme
  Classification with Application to Dynamic Search Advertising}. In
  \bibinfo{booktitle}{\emph{Proceedings of the 2018 World Wide Web Conference}}
  (Lyon, France) \emph{(\bibinfo{series}{WWW '18})}.
  \bibinfo{publisher}{International World Wide Web Conferences Steering
  Committee}, \bibinfo{address}{Republic and Canton of Geneva, CHE},
  \bibinfo{pages}{993–1002}.
\newblock
\showISBNx{9781450356398}
\urldef\tempurl%
\url{https://doi.org/10.1145/3178876.3185998}
\showDOI{\tempurl}


\bibitem[\protect\citeauthoryear{Prabhu, Kusupati, Gupta, and Varma}{Prabhu
  et~al\mbox{.}}{2020}]%
        {Prabhu2020ExtremeRF}
\bibfield{author}{\bibinfo{person}{Yashoteja Prabhu}, \bibinfo{person}{Aditya
  Kusupati}, \bibinfo{person}{Nilesh Gupta}, {and} \bibinfo{person}{Manik
  Varma}.} \bibinfo{year}{2020}\natexlab{}.
\newblock \showarticletitle{Extreme Regression for Dynamic Search Advertising}.
\newblock \bibinfo{journal}{\emph{Proceedings of the 13th International
  Conference on Web Search and Data Mining}} (\bibinfo{year}{2020}).
\newblock


\bibitem[\protect\citeauthoryear{Qi, Gong, Yan, Jiao, Shao, Zhang, Li, Duan,
  and Zhou}{Qi et~al\mbox{.}}{2020}]%
        {ProphetNetAds}
\bibfield{author}{\bibinfo{person}{Weizhen Qi}, \bibinfo{person}{Yeyun Gong},
  \bibinfo{person}{Yu Yan}, \bibinfo{person}{Jian Jiao}, \bibinfo{person}{B.
  Shao}, \bibinfo{person}{R. Zhang}, \bibinfo{person}{H. Li},
  \bibinfo{person}{N. Duan}, {and} \bibinfo{person}{M. Zhou}.}
  \bibinfo{year}{2020}\natexlab{}.
\newblock \showarticletitle{ProphetNet-Ads: A Looking Ahead Strategy for
  Generative Retrieval Models in Sponsored Search Engine}.
\newblock \bibinfo{journal}{\emph{ArXiv}}  \bibinfo{volume}{abs/2010.10789}
  (\bibinfo{year}{2020}).
\newblock


\bibitem[\protect\citeauthoryear{Qu, Ding, Liu, Liu, Ren, Zhao, Dong, Wu, and
  Wang}{Qu et~al\mbox{.}}{2021}]%
        {RocketQA}
\bibfield{author}{\bibinfo{person}{Yingqi Qu}, \bibinfo{person}{Yuchen Ding},
  \bibinfo{person}{Jing Liu}, \bibinfo{person}{Kai Liu},
  \bibinfo{person}{Ruiyang Ren}, \bibinfo{person}{Xin Zhao},
  \bibinfo{person}{Daxiang Dong}, \bibinfo{person}{Hua Wu}, {and}
  \bibinfo{person}{Haifeng Wang}.} \bibinfo{year}{2021}\natexlab{}.
\newblock \showarticletitle{RocketQA: An Optimized Training Approach to Dense
  Passage Retrieval for Open-Domain Question Answering}. In
  \bibinfo{booktitle}{\emph{NAACL}}.
\newblock


\bibitem[\protect\citeauthoryear{Ribeiro-Neto, Cristo, Golgher, and
  Moura}{Ribeiro-Neto et~al\mbox{.}}{2005}]%
        {RibeiroNeto2005ImpedanceCI}
\bibfield{author}{\bibinfo{person}{B. Ribeiro-Neto}, \bibinfo{person}{Marco
  Cristo}, \bibinfo{person}{P.~B. Golgher}, {and} \bibinfo{person}{E. Moura}.}
  \bibinfo{year}{2005}\natexlab{}.
\newblock \showarticletitle{Impedance coupling in content-targeted
  advertising}. In \bibinfo{booktitle}{\emph{SIGIR '05}}.
\newblock


\bibitem[\protect\citeauthoryear{Saini, Jain, Dave, Jiao, Singh, Zhang, and
  Varma}{Saini et~al\mbox{.}}{2021}]%
        {galaxc}
\bibfield{author}{\bibinfo{person}{Deepak Saini}, \bibinfo{person}{Arnav~Kumar
  Jain}, \bibinfo{person}{Kushal Dave}, \bibinfo{person}{Jian Jiao},
  \bibinfo{person}{Amit Singh}, \bibinfo{person}{Ruofei Zhang}, {and}
  \bibinfo{person}{Manik Varma}.} \bibinfo{year}{2021}\natexlab{}.
\newblock \showarticletitle{GalaXC: Graph Neural Networks with Labelwise
  Attention for Extreme Classification}.
\newblock \bibinfo{journal}{\emph{Proceedings of the Web Conference 2021}}
  (\bibinfo{year}{2021}).
\newblock


\bibitem[\protect\citeauthoryear{Sellam, Das, and Parikh}{Sellam
  et~al\mbox{.}}{2020}]%
        {bleurt}
\bibfield{author}{\bibinfo{person}{Thibault Sellam}, \bibinfo{person}{Dipanjan
  Das}, {and} \bibinfo{person}{Ankur~P. Parikh}.}
  \bibinfo{year}{2020}\natexlab{}.
\newblock \showarticletitle{BLEURT: Learning Robust Metrics for Text
  Generation}. In \bibinfo{booktitle}{\emph{Annual Meeting of the Association
  for Computational Linguistics}}.
\newblock


\bibitem[\protect\citeauthoryear{Vaswani, Shazeer, Parmar, Uszkoreit, Jones,
  Gomez, Kaiser, and Polosukhin}{Vaswani et~al\mbox{.}}{2017}]%
        {transformers}
\bibfield{author}{\bibinfo{person}{Ashish Vaswani}, \bibinfo{person}{Noam~M.
  Shazeer}, \bibinfo{person}{Niki Parmar}, \bibinfo{person}{Jakob Uszkoreit},
  \bibinfo{person}{Llion Jones}, \bibinfo{person}{Aidan~N. Gomez},
  \bibinfo{person}{Lukasz Kaiser}, {and} \bibinfo{person}{Illia Polosukhin}.}
  \bibinfo{year}{2017}\natexlab{}.
\newblock \showarticletitle{Attention is All you Need}.
\newblock \bibinfo{journal}{\emph{ArXiv}}  \bibinfo{volume}{abs/1706.03762}
  (\bibinfo{year}{2017}).
\newblock


\bibitem[\protect\citeauthoryear{Xiong, Xiong, Li, Tang, Liu, Bennett, Ahmed,
  and Overwijk}{Xiong et~al\mbox{.}}{2021}]%
        {ANCE}
\bibfield{author}{\bibinfo{person}{Lee Xiong}, \bibinfo{person}{Chenyan Xiong},
  \bibinfo{person}{Ye Li}, \bibinfo{person}{Kwok-Fung Tang},
  \bibinfo{person}{Jialin Liu}, \bibinfo{person}{Paul Bennett},
  \bibinfo{person}{Junaid Ahmed}, {and} \bibinfo{person}{Arnold Overwijk}.}
  \bibinfo{year}{2021}\natexlab{}.
\newblock \showarticletitle{Approximate Nearest Neighbor Negative Contrastive
  Learning for Dense Text Retrieval}.
\newblock \bibinfo{journal}{\emph{ArXiv}}  \bibinfo{volume}{abs/2007.00808}
  (\bibinfo{year}{2021}).
\newblock


\bibitem[\protect\citeauthoryear{Xue, Constant, Roberts, Kale, Al-Rfou,
  Siddhant, Barua, and Raffel}{Xue et~al\mbox{.}}{2020}]%
        {mt5}
\bibfield{author}{\bibinfo{person}{Linting Xue}, \bibinfo{person}{Noah
  Constant}, \bibinfo{person}{Adam Roberts}, \bibinfo{person}{Mihir Kale},
  \bibinfo{person}{Rami Al-Rfou}, \bibinfo{person}{Aditya Siddhant},
  \bibinfo{person}{Aditya Barua}, {and} \bibinfo{person}{Colin Raffel}.}
  \bibinfo{year}{2020}\natexlab{}.
\newblock \showarticletitle{mT5: A Massively Multilingual Pre-trained
  Text-to-Text Transformer}. In \bibinfo{booktitle}{\emph{North American
  Chapter of the Association for Computational Linguistics}}.
\newblock


\bibitem[\protect\citeauthoryear{Zhang, Kishore, Wu, Weinberger, and
  Artzi}{Zhang et~al\mbox{.}}{2019}]%
        {bert_score}
\bibfield{author}{\bibinfo{person}{Tianyi Zhang}, \bibinfo{person}{Varsha
  Kishore}, \bibinfo{person}{Felix Wu}, \bibinfo{person}{Kilian~Q. Weinberger},
  {and} \bibinfo{person}{Yoav Artzi}.} \bibinfo{year}{2019}\natexlab{}.
\newblock \showarticletitle{BERTScore: Evaluating Text Generation with BERT}.
\newblock \bibinfo{journal}{\emph{ArXiv}}  \bibinfo{volume}{abs/1904.09675}
  (\bibinfo{year}{2019}).
\newblock


\end{thebibliography}


 



\end{document}